\documentclass[conference]{IEEEtran}
\IEEEoverridecommandlockouts
\usepackage{cite}
\usepackage{amsmath,amssymb,amsfonts}
\usepackage{algorithmic}
\usepackage{graphicx}
\usepackage{textcomp}
\usepackage{xcolor}

\usepackage{booktabs}
\usepackage{longtable}

\newtheorem{example}{Example}

\newtheorem{proposition}{Proposition}
\newtheorem{definition}{Definition}

\newcommand{\tseitinvars}{\mathbf{X}}
\newcommand{\tseitinvar}{X}
\newcommand{\regvars}{\mathbf{Y}}

\newcommand{\circuit}{\psi}

\usepackage{mathtools}
\newcommand{\defeq}{\vcentcolon=}

\newcommand{\concept}[1]{#1}

\begin{document}

\title{Pruning Boolean d-DNNF Circuits Through Tseitin-Awareness
\thanks{This article is a longer version (includes supplementary material) of the one submitted to ICTAI2024.
This work was supported by the EU H2020 ICT48 project “TAILOR” under contract \#952215.}
}


\author{\IEEEauthorblockN{Vincent Derkinderen}
\IEEEauthorblockA{\textit{Dept. of Computer Science}, \textit{KU Leuven}, B-3000 Leuven, Belgium \\
}
}

\maketitle

\begin{abstract}
    Boolean circuits in d-DNNF form enable tractable probabilistic inference.
    However, as a key insight of this work, we show that commonly used d-DNNF compilation approaches introduce irrelevant subcircuits. We call these subcircuits Tseitin artifacts, as they are introduced due to the Tseitin transformation step -- a well-established procedure to transform any circuit into the CNF format required by several d-DNNF knowledge compilers. We discuss how to detect and remove both Tseitin variables and Tseitin artifacts, leading to more succinct circuits. We empirically observe an average size reduction of $77.5\%$ when removing both Tseitin variables and artifacts. The additional pruning of Tseitin artifacts reduces the size by $22.2\%$ on average. This significantly improves downstream tasks that benefit from a more succinct circuit, e.g., probabilistic inference tasks.
\end{abstract}


\section{Introduction}

\begin{figure}
    \centering
    \includegraphics{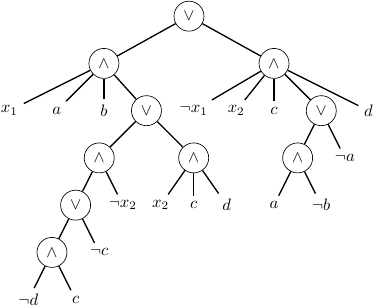}
    \caption{A d-DNNF representation of $\mathcal{T}(\psi)$, of Example~\ref{ex:tseitin}.}
    \label{fig:CDCL-full}
\end{figure}

Boolean circuits represent Boolean functions in a manner that allows certain operations to become feasible. In this work we focus on d-DNNF circuits (see Fig.~\ref{fig:CDCL-full} for an example), because they allow us to compute the weighted model count in polytime, despite it being a computationally hard task in general~\cite{Darwiche2002KCmap}. As such, they play an important role in enabling tractable probabilistic inference~\cite{ChaviraD08,suciu_probabilistic_2011,holtzen_scaling_2020,derkinderen2024semirings,ahmed2022pylon,cao2023scaling}. 
Furthermore, a key advantage of the circuit-approach to computing the weighted model count, is that when the weights change, the same compiled circuit can be reused and only the circuit evaluation must be repeated. This effectively amortizes the compilation cost across evaluations, and is especially relevant in the context of parameter learning where the parameters evolve over many evaluations.
However, in such a context, the circuit size also becomes important. In neuro-symbolic AI, for instance, where such circuits are being combined with neural networks, many evaluations take place during training and an efficient circuit evaluation is key to maintaining a fast enough neural network learning pipeline~\cite{ManhaeveDKDR21,XuZFLB18}. To conclude, besides efficiently obtaining a d-DNNF circuit, it is also important that these circuits are succinct.

To obtain a succinct d-DNNF circuit from any Boolean circuit, the field of knowledge compilation has developed several d-DNNF compilers. The CDCL-based class of compilers, which includes D4~\cite{Lagniez17D4} and sharpSAT-TD~\cite{KieselE23}, requires the input circuit to be in conjunctive normal form (CNF).
The most prominent algorithm to efficiently obtain this form for compilation is the \concept{Tseitin transformation}~\cite{Tseitin1983}. This is a procedure to convert any propositional Boolean circuit $\psi$ into an equisatisfiable version in CNF.

A side-effect of the Tseitin transformation, is that it introduces additional auxiliary variables. These variables can be removed after compilation through existential quantification~\cite{KieselE23}. However, as \textbf{our primary contribution}, we discuss the insight that this approach can result in tautological d-DNNF subcircuits. That is, these subcircuits are equivalent to true and can be removed entirely. We call these subcircuits \emph{Tseitin artifacts} as they emerge during compilation because of the prior Tseitin transformation. We discuss and formally define such artifacts in Section~\ref{sec:tseitin_artifacts}.

Tseitin artifacts can be efficiently detected within a compiled d-DNNF circuit by a single bottom-up pass of the representation. We discuss this approach, and when these artifacts are expected to emerge, in Section~\ref{sec:artifact_detection}, after which we briefly discuss related work in Section~\ref{sec:related_work}. Finally, in Section~\ref{sec:experiments}, we empirically confirm the prevalence of Tseitin artifacts and show that actively removing both Tseitin variables and Tseitin artifacts reduces the circuit size by an average of $77.5$\%. Compared to only removing Tseitin variables, the additional pruning of Tseitin artifacts reduces the size by $22.2\%$ on average ($29.7\%$ if we only consider the circuits where Tseitin artifacts were present).

\section{Background}

\subsection{Propositional Logic}

We use the standard terminology of propositional logic. A \concept{literal} is a Boolean variable $v$ or its negation $\neg v$. A \concept{propositional formula} $\psi$ is inductively defined as a literal $l$, the negation of a formula $\neg \psi_1$, a conjunction of formulas $\psi_1 \land \psi_2$, or a disjunction of formulas $\psi_1 \lor \psi_2$, each with the usual semantics.
We will also allow $\top$ (`true') and $\bot$ (`false') to appear in the formula, and when convenient, also the equivalence symbol $\psi_1 {\iff} \psi_2$ defined as $(\psi_1 \land \psi_2) \lor (\neg \psi_1 \land \neg \psi_2)$. A \concept{clause} is a literal or a disjunction of multiple literals, and a formula $\psi$ is in \concept{conjunctive normal form} (CNF) iff it is a clause or a conjunction of multiple clauses.

An \concept{interpretation} is a truth assignment to the Boolean variables. When formula $\psi$ is satisfied under the assignment, we say the interpretation is a \concept{model} of $\psi$. When a formula is trivially satisfied, that is, satisfied under any interpretation, we call it a \concept{tautology}. A trivial example of a tautology is $x \lor \neg x$.

We use $\psi|_{l}$ to denote formula $\psi$ conditioned on $l$ being true. This is different from $\psi \land l$, as $l$ and $\neg l$ no longer occur within $\psi|_l$.
The \concept{existential quantification} of a variable $X$ in $\psi$, denoted as $\exists X. \psi$, is semantically equivalent to $\psi|_x \lor \psi|_{\neg x}$. This process is also called ``forgetting variable $X$''. We extend this notation to operate on a set of variables $\mathbf{X} {=} \{X_1, X_2, \dots, X_n \}$: $\exists \mathbf{X}. \psi = \exists X_1 \exists X_2 \dots \exists X_n. \psi$.

\begin{example} Consider $\psi \defeq a \land (\neg b \lor c)$. This is a CNF formula of two clauses. There exist $2^3$ interpretations over $\mathbf{V} = \{A,B,C\}$, three of which are a model of $\psi$: $\{a, \neg b, c\}$, $\{a, \neg b, \neg c\}$ and $\{a, b, c\}$. $\psi$ is not a tautology. When conditioning on $a$, we have $\psi|_{a} = \neg b \lor c$.
\end{example}

\subsection{Weighted Model Counting}

\begin{definition}[weighted model count]
    Given a propositional formula $\psi$ over variables $\mathbf{V}$, and weight function $w$ that maps each literal to a real value, the weighted model count is defined as 
    \begin{equation}
        WMC(\psi, \mathbf{V}, w) = \sum_{\text{model of } \psi \;} \prod_{\text{lit} \in \text{model}} w(\text{lit}).
    \end{equation}
\end{definition}
We will use $WMC(\psi, w)$ when $\mathbf{V}$ is clear from context, and $MC(\psi)$ to refer to unweighted model counting, which is equivalent to using weight $1$ for all literals.
Since probabilistic inference can be cast as a weighted model counting (WMC) task~\cite{ChaviraD08,Fierens2015problog}, efficiently computing the WMC is of great interest. This is possible using a compilation-based approach.

\subsection{Circuit Properties}

The structure of a propositional formula $\psi$ is a tree. When allowing reuse of substructures, i.e., when we have a more general single-rooted DAG structure, we instead use the term Boolean circuit. There are several structural properties~\cite{ChaviraD08} that such a circuit may have, and that are of interest to our work.

\begin{definition}[determinism (d)]
    A Boolean circuit $\circuit$ is deterministic iff for each $\circuit_1 \lor \circuit_2$ within $\circuit$, $\circuit_1$ and $\circuit_2$ do not share any models. That is, $\circuit_1 \land \circuit_2 = \bot$.
\end{definition}
\begin{definition}[Decomposability (D)]
    A Boolean circuit $\circuit$ is decomposable iff for each $\circuit_1 \land \circuit_2$ within $\circuit$, $\circuit_1$ and $\circuit_2$ do not share any variables.
\end{definition}
\begin{definition}[Negation Normal Form (NNF)]
    A Boolean circuit is in negation normal form iff it consists only of $\lor$, $\land$, and literals (negation is only allowed over variables).
\end{definition}
\begin{definition}[smoothness (s)]
    A Boolean circuit $\circuit$ is smooth iff for each $\circuit_1 \lor \circuit_2$ within $\circuit$, $\circuit_1$ and $\circuit_2$ contain the same variables.
\end{definition}

When $\circuit$ satisfies the sd-DNNF properties, its WMC can be computed in time linear in the size of the representation by a single bottom-up evaluation.
The smoothness property is less important from a computational complexity perspective as it is obtainable in polytime while preserving the d-DNNF properties~\cite{darwiche2001tractable,Shih19Smoothing}.  
Furthermore, if $w(v) + w(\neg v) = 1$ holds for all variables $V$ (for example within a probabilistic inference context when the variables represent a Bernoulli distribution), then the smoothness property is unnecessary and a d-DNNF circuit suffices to efficiently compute the WMC.

\subsection{CDCL Algorithm}

Several algorithms have been developed to compile a Boolean circuit into a d-DNNF. Many state-of-the-art compilers are based on the CDCL algorithm that was initially developed for SAT solving, but can be adapted for compilation. Examples include D4~\cite{Lagniez17D4} and sharpsat-TD~\cite{Korhonen23sharpsatTD,KieselE23}. 
For a more detailed explanation of these algorithms we refer to \cite{HuangD05,Lagniez17D4,KieselE23}.

Important to understanding our contributions is that these algorithms operate on a CNF formula $\psi$, and that they iteratively condition on literals until the remaining formula is satisfied.

\begin{example} Consider the following $\psi$, a CNF of two clauses.
    \begin{equation}
        \psi \defeq (a \lor \neg b) \land (a \lor \neg c)
    \end{equation}
    If a CDCL-based algorithm were to condition on $\neg b$, clauses that contain $\neg b$ are removed, and $b$ is removed from every clause. The remaining formula is $\psi|_{\neg b} = (a \lor \neg c)$.
    When then conditioning on $a$, the remaining formula is $\psi|_{\neg b,a} = \top$, and the algorithm backtracks to cover the other models in a similar fashion.
\end{example}

Also relevant is component decompositioning~\cite{Pehoushek00}, a very effective optimisation used within weighted model counting and d-DNNF compilation to reduce the search space. This optimisation exploits the fact that if a CNF can be partitioned into sets of clauses that do not share any variables (called components), then these can be treated separately. Furthermore, these components can be cached such that when they are encountered again later in the search space, the cached result can be reused. For d-DNNF compilation, this enables reuse of subcircuits and results in a DAG rather than a tree. The effectiveness of this optimisation has lead the community to propose a variety of component representations~\cite{Thurley06,SharmaRSM19,van2021symmetric}.

\begin{example} Consider the following CNF of three clauses.
    \begin{equation*}
        \psi \defeq (a \lor \neg b) \land (a \lor \neg c) \land (\neg d \lor e)
    \end{equation*}
    The first two clauses together form a component $\mathcal{C}_1$, while the third clause forms component $\mathcal{C}_2$ as it does not share any variables with the previous clauses. From a weighted model count perspective we have \eqref{eq:ex:wmc} which indicates that the model count of each component can be computed separately.
    \begin{equation}\label{eq:ex:wmc}
        WMC(\psi,\mathbf{V},w) = WMC(\mathcal{C}_1,w) \times
            WMC(\mathcal{C}_2,w),
    \end{equation}
    Similarly for d-DNNF compilation, both components can be compiled separately and combined using a conjunction $\land$.
\end{example}

CDCL-based algorithms require the input $\psi$ to be a CNF formula, which is obtainable using the Tseitin transformation.

\subsection{Tseitin Transformation}\label{sec:bg:tseitin}

The Tseitin (or Tseytin) transformation is a procedure by which any Boolean circuit $\psi$ can be converted into CNF~\cite{Tseitin1983}.
The main idea powering this transformation is that, when we introduce a new auxiliary Boolean variable $x$ to refer to a subcircuit $\psi'$, we can conveniently replace each occurrence of $\psi'$ by $x$. For larger nested circuits this significantly simplifies the CNF transformation because each $\big(l \Leftrightarrow \bigvee_i l_i\big)$ and $\big(l \Leftrightarrow \bigwedge_i l_i\big)$ can easily be transformed into a CNF where the number of clauses is exactly equal to the number of literals $l_i$ and $l$. The number of clauses in the resulting CNF is therefore linear in the number of subcircuits $\psi'$ and in the number of literals occurring in $\psi'$.

\begin{example}[Tseitin transformation]\label{ex:tseitin}
Consider the circuit $\psi$ given below.
\begin{equation}\label{eq:ex_psi}
    \psi \defeq \quad (a \land b) \vee (c \land d)
\end{equation}
The Tseitin transformation introduces new variables $x_1$ and $x_2$ to refer to $(a \land b)$ and $(c \land d)$ respectively. Using this new equivalence, the original circuit $\psi$ could be summarized as
\begin{equation}\label{eq:ex_psi_tseitin}
    \mathcal{T}(\psi) \defeq 
    \big(x_1 \lor x_2 \big) \land 
    \big(x_1 {\iff} a \land b \big) \land 
    \big(x_2 {\iff} c \land d \big).
\end{equation}
As CNF, this results in circuit $\text{CNF}(\mathcal{T}(\psi))$, displayed below as a conjunction of seven clauses.
\begin{align}\label{eq:ex_psi_tseitin_cnf}
    \big(& x_1 \lor \neg a \lor \neg b \big) \land
    \big( \neg x_1 \lor a \big) \land (\neg x_1 \lor b) \land \\
    \big(& x_2 \lor \neg d \lor \neg c \big) \land
    \big(\neg x_2 \lor d \big) \land
    \big(\neg x_2 \lor c \big) \land \big(x_1 \lor x_2) \nonumber
\end{align}
\end{example}
In this example we distinguished $\mathcal{T}(\psi)$ from $\text{CNF}(\mathcal{T}(\psi))$. In the remainder of this paper we use $\mathcal{T}(\psi)$ to refer to the circuit resulting from the Tseitin transformation on $\psi$. Its exact representation, as a CNF~\eqref{eq:ex_psi_tseitin_cnf} or not~\eqref{eq:ex_psi_tseitin}, will generally not be important. In case it is important, the representation we refer to will be clear from context.

Note that $\psi$ and $\mathcal{T}(\psi)$ are not equivalent as $\mathcal{T}(\psi)$ contains variables that are not present in $\psi$, and that were introduced by the Tseitin transformation. We call these \concept{Tseitin variables}. Importantly, there is a one-to-one mapping between the models of $\psi$ and the models of $\mathcal{T}(\psi)$~\cite{Oztok2017Forgetting}. By construction, each model of $\psi$ implies a truth value for the Tseitin variables in $\mathcal{T}(\psi)$ through the equivalences introduced by the transformation~\eqref{eq:ex_psi_tseitin}. In the other direction, ignoring the Tseitin variables from a model of $\mathcal{T}(\psi)$ yields exactly one model of $\psi$. As a consequence, we have the following relation \eqref{eq:tseitin_existential_quantification} where $\mathbf{X}$ is the set of Tseitin variables introduced during the Tseitin transformation.
\begin{equation}\label{eq:tseitin_existential_quantification}
    \exists \mathbf{X}. \mathcal{T}(\psi) = \psi
\end{equation}

When computing the weighted model count of $\psi$ using $\mathcal{T}(\psi)$, the weight of a Tseitin variable $X$ is typically set to one ($w(x) {=} 1 {=} w(\neg x)$). This preserves the weight of each model and consequently also the weighted model count of $\psi$.

\section{What are Tseitin Artifacts?}\label{sec:tseitin_artifacts}

We first informally introduce the concept that we refer to as a \emph{Tseitin artifact}.
A more formal definition is then provided afterwards.
We study these artifacts in the context of d-DNNF compilation. Therefore, suppose we wish to obtain a d-DNNF representation of circuit $\psi$ \eqref{eq:ex_psi} in Example~\ref{ex:tseitin}. We will use a CDCL-based d-DNNF compiler that requires a CNF as input, so we first use the Tseitin transformation to obtain $\mathcal{T}(\psi)$ \eqref{eq:ex_psi_tseitin_cnf}. Fig.~\ref{fig:CDCL-full} shows a possible output of the compiler, a d-DNNF circuit representing $\mathcal{T}(\psi)$. Since we are interested in the number of computations during evaluation, we define the circuit size as the number of binary nodes (e.g., a node with three inputs represents two binary nodes, and so on). The size of the d-DNNF in Fig.~\ref{fig:CDCL-full} is $16$.

\subsection{Existential Quantification of Tseitin Variables}\label{subsec:existential_quantification}

The circuit in Fig.~\ref{fig:CDCL-full} is a d-DNNF and can hence be used to efficiently compute the weighted model count of $\mathcal{T}(\psi)$ and $\psi$.
The current representation, however, does contain many unnecessary elements. In particular, the Tseitin variables $\tseitinvars$ were introduced by the Tseitin transformation to easily obtain a CNF, but are irrelevant to downstream counting tasks.
These variables can be removed through existential quantification, i.e., by computing $\exists \tseitinvars. \mathcal{T}(\psi)$.

Thanks to the decomposability property, existential quantification of $\tseitinvar \in \tseitinvars$ is very simple here: replace each occurrence of $x$ and $\neg x$ in the formula by $\top$. We refer to this procedure as $\textsc{EXISTS}(\psi,X)$. This procedure is guaranteed to preserve the DNNF properties~\cite{Oztok2017Forgetting}. Additionally, recent work has proven that since we existentially quantify Tseitin variables, this procedure is guaranteed to also preserve determinism~\cite{KieselE23}.

\begin{figure}
    \centering
    \includegraphics{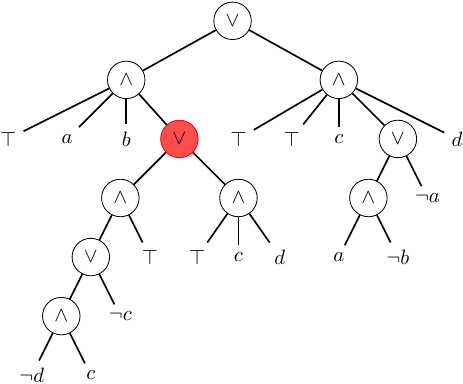}
    \caption{The d-DNNF of Fig.~\ref{fig:CDCL-full}, with the Tseitin variables $x_1$ and $x_2$ existentially quantified (that is, $\exists x_1,x_2. \mathcal{T}(\psi)$). The red node indicates the root of a subcircuit that is a Tseitin artifact.}
    \label{fig:CDCL-replaced}
\end{figure}

Fig.~\ref{fig:CDCL-replaced} shows the circuit of Fig.~\ref{fig:CDCL-full} after performing existential quantification of the Tseitin variables. Combining the results above with \eqref{eq:tseitin_existential_quantification}, it follows that Fig.~\ref{fig:CDCL-replaced} again represents circuit $\psi$ instead of $\mathcal{T}(\psi)$, but now as a d-DNNF due to the earlier compilation step.

The $\top$ values in Fig.~\ref{fig:CDCL-replaced} can be propagated\footnote{When we refer to the $\textsc{EXISTS}$ procedure, we include the propagation step}, simplifying occurrences of the form $\psi_i \lor \top$ and $\psi_i \land \top$. This reduces the circuit size to $11$. 
Importantly, we show in the next section that the circuit size can be reduced even more.

\subsection{Tseitin Artifacts}\label{subsec:tseitin_artifcats}

The subcircuit $f$ whose root is marked in red in Fig.~\ref{fig:CDCL-replaced}, is given in \eqref{eq:subformula_f}.
While this subcircuit initially represented $\big( x_2 {\iff} c \land d \big)$, which is not a tautology, after performing the existential quantification of $x_2$, it did become a tautology.
\begin{equation}\label{eq:subformula_f}
    \exists x_2. f \equiv \Big(\big((\neg d \land c) \lor \neg c\big) \land \top \Big) \lor \Big( \top \land c \land d \Big) \equiv \top
\end{equation}
This means that $f$ can be entirely replaced by $\top$. Indeed, consider the initial circuit $\psi$ of Example~\ref{ex:tseitin} and note that $\psi|_{a,b} \equiv \top$. We call $f$ a Tseitin artifact, a tautological subcircuit introduced to the compiled circuit due to the Tseitin transformation.

\begin{definition}[Tseitin artifact]
    A (sub)circuit $f(\tseitinvars, \regvars)$ over Tseitin variables $\tseitinvars$ and non-Tseitin variables $\regvars$ is a Tseitin artifact if and only if it is a tautology when existiantially quantifying over $\tseitinvars$. That is, when $\exists \tseitinvars. f(\tseitinvars,\regvars) \equiv \top$. 
\end{definition}

This definition can be generalised beyond Tseitin variables $\tseitinvars$, to variables that are completely defined by other variables (cf. the definition of \emph{definability} by~\cite{Lagniez16definability,KieselE23}).
Our focus on Tseitin variables is a practical choice. First, they are known to be defined by other variables so we do not need a procedure to determine the set of defined variables. Second, they are irrelevant from a user perspective. That is, these variables are not present in the original circuit (pre-Tseitin transformation) so we can safely assume they are irrelevant to any downstream model counting task and can thus be removed.

Our insight can be applied more generally: if a (sub)circuit is equivalent to $\top$, you can replace it with $\top$ to reduce the circuit size. However, a top-down d-DNNF compiler would never produce such subcircuits under normal conditions because a non-empty CNF is never equivalent to $\top$ (unless we allow CNFs of only trivial clauses such as $x \lor \neg x$). Subcircuits that are equivalent to $\top$ only emerge within a d-DNNF once we perform existential quantification, which is advised when using the Tseitin transformation as we show in the experiments. We therefore position our insight around Tseitin variables, and refer to the subcircuits as Tseitin artifacts.

Existential quantification of Tseitin variables using the $\textsc{EXISTS}$ procedure of section~\ref{subsec:existential_quantification} does not eliminate all Tseitin artifacts. For example, while propagating $\top$ in Fig.~\ref{fig:CDCL-replaced}, we would not have realised that subcircuit $f$ (node marked in red) is equivalent to $\top$, missing the possible reduction to a circuit of size $6$ instead of $11$. In the next section we study how to detect these artifacts and how they emerge.

\section{Detecting Tseitin Artifacts}\label{sec:artifact_detection}

\subsection{How To Detect Them?}
Tseitin artifacts can be detected in an sd-DNNF representation in time linear in the size of the sd-DNNF.

\begin{proposition}\label{th:artifact_mc}
    When $f(\tseitinvars, \regvars)$ is a (sub)circuit of a d-DNNF representation for $\mathcal{T}(\psi)$, over Tseitin variables $\tseitinvars$ and non-Tseitin variables $\regvars$, then $f$ is a Tseitin artifact if and only if the unweighted model count is $MC(f) = 2^{|\regvars|}$.
\end{proposition}
\begin{IEEEproof}
    We know $MC(f) = MC(\exists \tseitinvars. f(\tseitinvars, \regvars))$, because by construction a Tseitin variable $\tseitinvar$ is completely defined by non-Tseitin variables $\regvars$.
    \begin{itemize}
        \item[] $MC(f) = 2^{|\regvars|} \iff MC(\exists \tseitinvars. f(\tseitinvars, \regvars)) = 2^{|\regvars|} \iff$
        \item[] $\exists \tseitinvars. f(\tseitinvars, \regvars) = \top \iff f(\tseitinvars, \regvars) \text{ is a Tseitin artifact}$ 
    \end{itemize}
\end{IEEEproof}

It is well known that the model count for each sd-DNNF (sub)circuit can be computed in time linear in the size of the representation, and that a similar approach can be used for non-smooth d-DNNF representations by performing the appropriate smoothing operations in polytime either before or during evaluation~\cite{Darwiche2002KCmap,Shih19Smoothing}. Furthermore, since we only require unweighted model counts, smoothing is unnecessary and we only need to know the number of variables that would have to be smoothed over. The number of these so called free variables is easy to extract during the compilation process.
Combining this fact with Proposition~\ref{th:artifact_mc}, we can detect Tseitin artifacts by computing the unweighted model count of each subcircuit $f(\tseitinvars, \regvars)$ using a single bottom-up evaluation. If the model count is equivalent to $2^{|\regvars|}$, $f$ is a Tseitin artifact and it can be removed when we existentially quantify over $\mathbf{X}$.

\subsection{When Do They Emerge?}\label{subsec:when_do_they_emerge}
Next we provide more intuition on when these artifacts may emerge within CDCL-based d-DNNF compilers. This helps to identify the types of circuits for which our proposed technique (removing Tseitin artifacts) has a high impact.
Consider again $\psi$ and $\mathcal{T}(\psi)$ from Example~\ref{ex:tseitin}. 
\begin{align*}
    \psi &\defeq (a \land b) \lor (c \land d) \\
    \mathcal{T}(\psi) &\defeq (x_1 \lor x_2) \land (x_1 {\iff} a \land b) \land (x_2 {\iff} c \land d)
\end{align*}
When a d-DNNF compiler conditions on $x_1$~\eqref{eq:tseitin_example_conditioned}, the clause $(x_1 \lor x_2)$ becomes satisfied and, since $x_2$ is not used elsewhere, the Tseitin equivalence $(x_2 {\iff} c \land d)$ will emerge as a Tseitin artifact (after component decompositioning which splits off the $(a \land b)$ part).
\begin{equation}\label{eq:tseitin_example_conditioned}
    \mathcal{T}(\psi)|_{x_1} = (a \land b) \land (x_2 {\iff} c \land d)
\end{equation}

\begin{figure}
    \centering
    \includegraphics{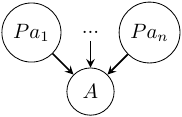}
    \caption{The noisy-OR BN where $Prob(pa) = \theta_{pa}$. When $Pa$ is true and its signal is not noisy ($\theta_{pa}^a$), $A$ becomes true.}
    \label{fig:noisy_or_BN}
\end{figure}
\begin{figure}
    \centering
    \includegraphics[width=\linewidth]{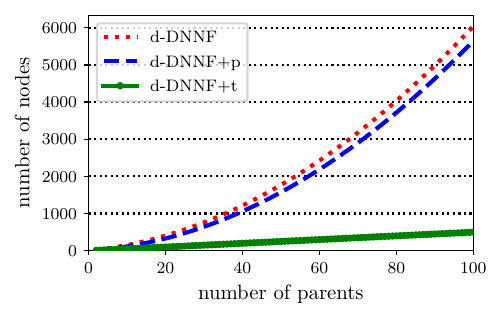}
    \caption{The d-DNNF circuit size for varying sizes of the noisy-OR problem. d-DNNF+p represents $\textsc{EXISTS}(\mathcal{T}(\psi), \mathbf{X})$. d-DNNF+t represents the additional removal of Tseitin artifacts. Lower is better.}
    \label{fig:noisy_or_results}
\end{figure}
More generally, if $\psi_1$ and $\psi_2$ are circuits more complex than a single literal, appearing together as $\psi_1 \lor \psi_2$, such that $\psi_1 \land \psi_2 \neq \bot$, then a Tseitin artifact may emerge depending on the variable ordering of the CDCL compiler. We empirically confirm this using a circuit that represents a noisy-or Bayesian network~\cite{meert2016relaxed}, which has a structure similar to Example~\ref{ex:tseitin}. The circuit $\psi$ is given in the equation below, while the Bayesian network is illustrated in Fig.~\ref{fig:noisy_or_BN}.
\begin{equation}\label{eq:noisy_or_psi}
    \begin{split}
    \psi &\defeq a \land \big(a \iff \bigvee_{Pa \in \mathbf{Pa}(A)} (\theta_{pa} \land \theta^a_{pa}) \big) \\
    \mathcal{T}(\psi) &\defeq a \land \big(a \iff \bigvee_{Pa \in \mathbf{Pa}(A)} x_{pa}\big) \land \\
    &\qquad \bigwedge_{Pa \in \mathbf{Pa}(A)} (x_{pa} \iff \theta_{pa} \land \theta^a_{pa}).
    \end{split}
\end{equation}
Fig.~\ref{fig:noisy_or_results} shows the results, confirming that the number of Tseitin artifacts increases as the circuit grows, and that removing them significantly reduces the circuit size (d-DNNF+t). Additionally, it shows that the $\textsc{EXISTS}$ procedure on its own also reduces the formula size but not nearly as much (d-DNNF+p).


Note the previously stated condition of $\psi_1 \land \psi_2 \neq \bot$. If $\psi_1$ and $\psi_2$ were to be mutually exclusive ($\psi_1 \land \psi_2 = \bot$) and not complete ($\psi_1 \lor \psi_2 \neq \top$), then resolving $\psi_1$ would imply $\neg \psi_2$, reducing the chance of a Tseitin artifact.
We empirically confirmed this using $8$ Bayesian networks whose conditional probability tables are encoded in a way that each $\lor$ is mutually exclusive~\cite{bnlearn}. A limited number of artifacts did emerge for two instances (see Appendix~\ref{app:andes} for a brief explnation), but overall there was little difference between existential quantification (d-DNNF+p) and the removal of Tseitin artifacts (d-DNNF+t).  We conclude that under these conditions artifacts may emerge, but it becomes more dependent on the variable ordering of the d-DNNF compiler and more unlikely.


Both the circuit $\psi$ and the variable ordering within the d-DNNF compiler influence the prevalence of Tseitin artifacts. We illustrate the influence of the latter using Example~\ref{ex:var_order}.
\begin{example}[variable order impact]\label{ex:var_order} Suppose we have
    \begin{equation}
        \begin{split}
            \psi &\defeq (a \land b) \lor c \\
            \mathcal{T}(\psi) &\defeq (x_1 \lor c) \land (x_1 {\iff} a \land b).
        \end{split}
    \end{equation}
    If the d-DNNF compiler first conditions on $c$, a Tseitin artifact emerges. If it instead conditions on $x_1$, or on $a$ and $b$, no Tseitin artifact emerges.
\end{example}

\section{Related Work}~\label{sec:related_work}

Prior work \cite{Oztok2017Forgetting} has studied the usage of existential quantification, the EXISTS procedure, to obtain succinct DNNF circuits. Continuing on this result, \cite{KieselE23} proved that existential quantification of $X$ also preserves determinism if $X$ is defined in terms of the other variables, and is thus applicable when targetting d-DNNF circuits. Our work continues on these findings, realising that this procedure may result in subcircuits that can be removed to reduce the circuit size even further.

The concept of definability, which strongly relates to the Tseitin variables, has previously been used in preprocessing, altering the CNF to improve model counting~\cite{Lagniez16definability,Lagniez20definability}. 
Similarly, the variable elimination approach of \cite{KieselE23} could eliminate Tseitin variables prior to compilation. While this is beneficial in some cases, it may also degrade performance in the context of Tseitin variables as these were introduced exactly to ensure a small CNF size (with likely faster compilation)~\cite{KieselE23}. For this reason they consider heuristics to determine which variables to eliminate. Furthermore, prior elimination of the Tseitin variables prevents the compiler from conditioning on them. This may lead to larger circuit sizes and a potential increase in compilation time.

\cite{Dubray23} proposed a novel projected model counter restricted to Horn clauses, which are clauses whose form is equivalent to $(\bigwedge_i l_i) \Rightarrow h$, with $l_i$ and $h$ positive literals or $\top$. The relation to our work is their propagation technique. This technique is based on the insight that during the counting process, if $h$ is an auxiliary variable that is not constrained by any other remaining Horn clause, then its associated clause can be satisfied without impacting the count. In other words, they can remove such clauses (only because $h$ is an auxiliary variable). This relates to the Tseitin artifacts, although these emerge from an equivalence structure rather than an implication and are not restricted to projected model counting.

\section{Experiments}\label{sec:experiments}

Given the influence of the variable ordering and the structure of disjunctions within $\psi$, a natural research question is: how prevalent are Tseitin artifacts? We primarily focus on the former, avoiding instances that we know will not contain Tseitin artifacts (i.e., Bayesian networks with mutual exclusive disjunctions). We study the effect of removing Tseitin artifacts on the circuit size, and compare the effect of performing simple existential quantification (denoted as d-DNNF+p) versus removing the Tseitin artifacts entirely as well (denoted as d-DNNF+t).

\subsection{Datasets}

\paragraph{Reverse-engineered CNFs (MCC)} Benchmarks used for weighted model counting and d-DNNF compilation are typically CNF formulas already, so these are not useful from a Tseitin transformation perspective. In our experiments however, we will assume these CNFs were produced by a Tseitin transformation before they were made publicly available. We determine the set of Tseitin variables $\mathbf{X}$ by identifying equivalences ($l \Leftrightarrow \bigvee_i l_i$ or $l \Leftrightarrow \bigwedge_i l_i$). As dataset, we consider the CNF instances of the 2022 and 2023 model counting competitions (MCC)~\cite{MCC2022,MCC2023}, only using instances with more than $25\%$ Tseitin variables. We ran the D4 d-DNNF compiler~\cite{Lagniez17D4} with a timeout of 3600 seconds, resulting in $65$ completed instances (after removing duplicate instances).

\paragraph{CNFs with Tseitin variables (CNFT)} We also consider the dataset of \cite{KieselE23}: ``a new set of benchmarks using two tools to translate (probabilistic) logic programs to CNFs~\cite{JanhunenN11,EiterHK21} on standard benchmarks from probabilistic logic programming''. By construction, their auxiliary variables are completely defined in terms of the other remaining variables. We include additional programs that were translated using ProbLog~\cite{Fierens2015problog}, for instance the power transmission networks from \cite{LatourBN19}, originating from \cite{wiegmans2016}. These two datasets resulted in $146$ and $34$ instances respectively (already excluding timeouts). From the neuro-symbolic (NeSy) AI setting we include the Countries knowledge graph: ``Countries is a knowledge graph with the $locatedIn$ and $neighborOf$ relations between the countries and continents on Earth''~\cite{MaeneR23}. There are three tasks leading to three DNF formulas. While more instances can be created, the results are expected to be similar since the structure will be similar. We therefore consider the three DNF formulas, and apply the Tseitin transformation to obtain CNFs and Tseitin variables $\mathbf{X}$.

\subsection{Results}

\begin{figure}
    \centering
    \includegraphics{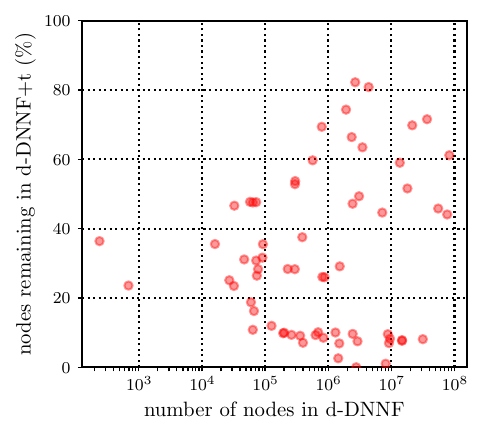}
    \caption{The fraction of d-DNNF nodes remaining after removing Tseitin artifacts (65 instances of the MCC dataset). Lower is better.}
    \label{fig:mcc_reg_vs_tseitin}
\end{figure}
\begin{figure}
     \centering
     \includegraphics{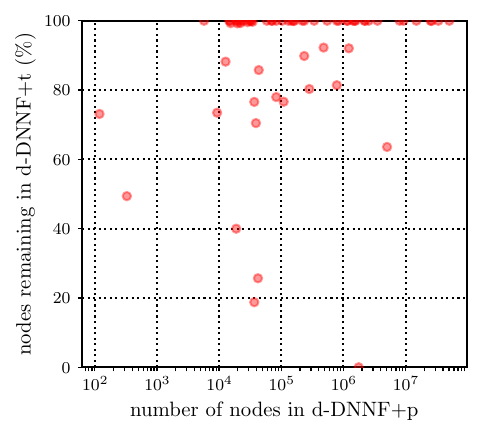}
    \caption{The fraction of d-DNNF nodes remaining after removing Tseitin artifacts, compared to only doing existential quantification. Lower is better.}
     \label{fig:mcc_prop_vs_tseitin}
 \end{figure}

\paragraph{MCC dataset} Fig.~\ref{fig:mcc_reg_vs_tseitin} shows the relative number of nodes that remain after removing the Tseitin artifacts and performing existential quantification. On average, this prunes $68.3\%$ of the nodes (std.\ of $22.6\%$). The additional contribution of detecting Tseitin artifacts, that is, d-DNNF+p compared to d-DNNF+t, is shown in Fig.~\ref{fig:mcc_prop_vs_tseitin}. This improves $27$ out of the $65$ instances, with an average of $24.0\%$ nodes pruned (for those $27$, with std.\ of $26.7\%$).

\begin{figure}
    \centering
    \includegraphics[width=\linewidth]{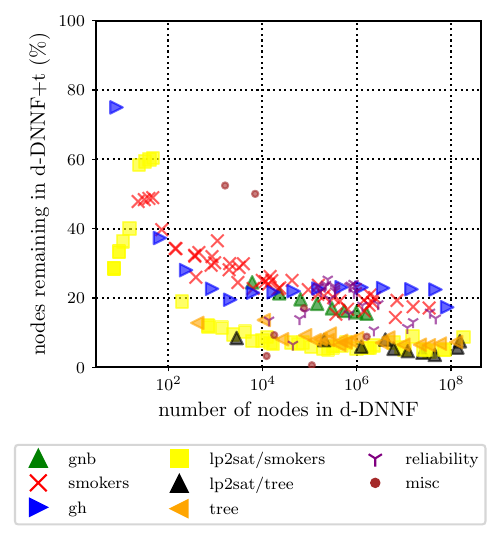}
    \caption{The fraction of d-DNNF nodes remaining after removing Tseitin artifacts (180 instances of the CNFT dataset). Lower is better. The different markers indicate the different instance classes within the dataset.}
    \label{fig:cnft_reg_vs_tseitin}
\end{figure}

\begin{figure}
     \centering
     \includegraphics[width=\linewidth, trim=0 2cm 0 0, clip]{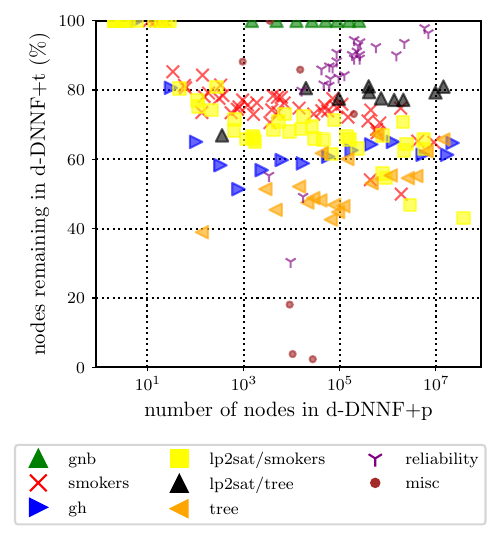}
    \caption{The fraction of d-DNNF nodes remaining after removing Tseitin artifacts (180 instances of the CNFT dataset), compared to only doing existential quantification. Lower is better.}
     \label{fig:cnft_prop_vs_tseitin}
 \end{figure}

\paragraph{CNFT dataset} Fig.~\ref{fig:cnft_reg_vs_tseitin} shows the relative number of nodes that remain after removing the Tseitin artifacts and performing existential quantification. On average, this prunes $81.1\%$ of the nodes (std.\ of $13.1\%$). The additional contribution of detecting Tseitin artifacts, that is, d-DNNF+p compared to d-DNNF+t, is shown in Fig.~\ref{fig:cnft_prop_vs_tseitin}. This improves $155$ out of the $180$ instances, with an average of $30.6\%$ nodes pruned (for those $155$, with std.\ of $15.7\%$). The three NeSy instances we illustrate separately, in Table~\ref{tab:nesy}. 


\begin{table}
\centering
\caption{The NeSy dataset instances: the original d-DNNF size, after performing existential quantification, and after additionally removing the Tseitin artifacts. The included percentages show the number of nodes relative to the original d-DNNF.}
\begin{tabular}{lrrr}
\toprule
inst  & $|\text{d-DNNF}|$ & $|\text{d-DNNF+p}|$ & $|\text{d-DNNF+t}|$ \\
\midrule
S1 & 26 268 407  & 16 828 763 (64\%) & 13 090 502 (50\%) \\
S2 & 83 265 347  & 62 161 281 (75\%) & 36 721 586 (44\%) \\
S3 & 79 311 007  & 33 708 205 (43\%) & 16 714 673 (21\%) \\
\end{tabular}
\label{tab:nesy}
\end{table}

\paragraph{Discussion} The overall results indicate the importance of existentially quantifying over irrelevant variables, and the additional reductions that can be achieved through the removal of Tseitin artifacts. 
As previously discussed, Tseitin artifacts do not necessarily appear in every circuit. This is observed in Fig.~\ref{fig:cnft_prop_vs_tseitin} where the circuits of class \emph{gnb} do not result in any additional reduction, meaning no Tseitin artifacts arose. We hypothesize, and manually confirmed for a few disjunctions, that the disjunctions within the \emph{gnb} circuits are mutually exclusive. This supports the explanation in Section~\ref{subsec:when_do_they_emerge}: disjunctions that do overlap are likely to result in Tseitin artifacts, while those that are mutually exclusive rarely lead to an artifact. For the remaining circuits the benefit is very clear, especially considering the low-cost of existential quantification and detection of Tseitin artifacts.

\section{Conclusion}

The Tseitin transformation introduces auxiliary variables to obtain a small CNF circuit. These variables can easily be removed after d-DNNF compilation. But as we have shown, subcircuits that are trivially satisfied may then emerge. Fortunately, we can easily detect and remove such artifacts using a single bottom-up evaluation of the d-DNNF circuit. We have empirically shown the positive impact of pruning these artifacts on the final circuit size. 
In future work, we will investigate the detection and removal of these artifacts during compilation, with the additional aim of reducing compilation time. This requires a new detection mechanism as the current one relies on each subcircuit to already be compiled.

\section*{Acknowledgment}
The author thanks Wannes Meert for the valuable discussions on this topic, and Jaron Maene for providing the knowledge graph DNF formulas.

\appendices

\section{Mutually Exclusive Disjunction}\label{app:andes}

\begin{figure}
    \centering
    \includegraphics{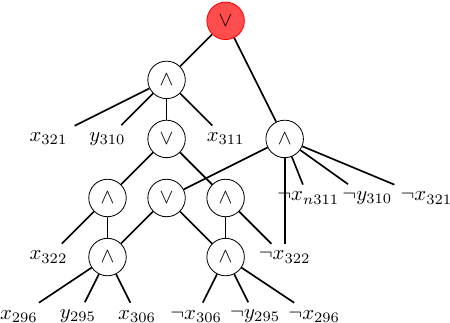}
    \caption{Example of a Tseitin artifact within the andes formula instance.}
    \label{fig:andes_artifact}
\end{figure}

A Bayesian network whose conditional probability tables are encoded in a way that each $\lor$ is mutually exclusive, results in few Tseitin artifacts. We illustrate one such probability table, $P(C|A,B)$:
\begin{equation}\label{eq:app:bn_cpt}
    c {\iff} \begin{cases}
        \big( a \land b \land \theta_{a,b}^c \big) \lor \\
        \big( a \land \neg b \land \theta_{a,\neg b}^c \big) \lor \\
        \big( \neg a \land b \land \theta_{\neg a,b}^c \big) \lor \\
        \big( \neg a \land \neg b \land \theta_{\neg a,\neg b}^c \big).
    \end{cases}
\end{equation}
We now briefly discuss one of the few found Tseitin artifacts. The subformula that formed this artifact was (using equivalences instead of CNF for clarity, and $x$ and $y$ to distinguish the Tseitin and non-Tseitin variables respectively):
\begin{equation}\label{eq:app:andes_explained}
\begin{split}
 (x_{306} &{\iff} x_{296}) \land (x_{296} {\iff} y_{295}) \land \\
 (x_{321} &{\iff} x_{311}) \land (x_{311} {\iff} y_{310}) \land \\
 (x_{322} &{\iff} x_{321} \land x_{306}).
\end{split}
\end{equation}
The Tseitin artifact of this formula is illustrated in Fig.~\ref{fig:andes_artifact}. To understand how this artifact emerged, we explain the meaning behind \eqref{eq:app:andes_explained} and how the conditioning process led to its existence.
The role of $x_{306}$ within andes is akin to $c$ in \eqref{eq:app:bn_cpt}. $(x_{306} {\iff} x_{296})$ was a larger equivalence with a four-case disjunction, similar to how \eqref{eq:app:bn_cpt} has four cases. After conditioning on several literals during the compilation process, the equivalence was reduced to $(x_{306} {\iff} x_{296})$. The case itself, like $a \land b \land \theta_{a,b}^c$, is represented by $x_{296}$, whose definition was reduced to $x_{296} {\iff} y_{295}$ (where $y_{295}$ is akin to $\theta_{a,b}^c$). The third and fourth conjunct within \eqref{eq:app:andes_explained} are explained analogously to the first two. Finally, $x_{306}$ and $x_{321}$ were both used within the case of another variable that depended on them (represented by $x_{322}$). This formed the final conjunct, that prevented their equivalences from being compiled independently.

\bibliographystyle{IEEEtran}
\bibliography{ictai2024/bibfile}

\newpage
\onecolumn
\section{Empirical results}

\begin{figure}
    \centering
    \includegraphics[width=0.5\linewidth]{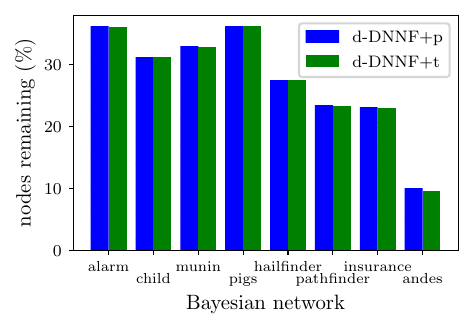}
    \caption{The number of nodes remaining after applying both procedures (y-axis) on a few formulas (x-axis). Each formula represents a query in a Bayesian network. d-DNNF+p represents the procedure of $\textsc{EXISTS}(\mathcal{T}(\psi), \mathbf{X})$. d-DNNF+t represents the additional removal of Tseitin artifacts. Lower is better.}
    \label{fig:nb_no_tseitin_results}
\end{figure}

Table~\ref{resultstable}~and~\ref{resultstable2} report the empirical results in more detail. Fig.~\ref{fig:nb_no_tseitin_results} shows the results of the Bayesian network experiments.

\begin{longtable}{|l|r|r|r|r|r|}
\caption{Experimental results comparing the compiled formula size ($|$d-DNNF$|$) to the size after performing existential quantification ($|$d-DNNF+p$|$) and to the size after also considering Tseitin artifacts ($|$d-DNNF+t$|$). The percentages are relative to the d-DNNF size. The percentage next to $|$Tseitin Vars$|$ indicates the relative number of tseitin variables compared to the total number of variables. The instances are those of the CNFT dataset, excluding the NeSy DNF formulas.}\label{resultstable}\\
\hline
\textbf{Benchmark} & \textbf{$|$Vars$|$} & \textbf{$|$Tseitin Vars$|$} & \textbf{$|$d-DNNF$|$} & \textbf{$|$d-DNNF+p$|$} & \textbf{$|$d-DNNF+t$|$} \\
\hline
\endfirsthead
\multicolumn{6}{c}%
{\tablename\ \thetable\ -- \textit{Continued from previous page}} \\
\hline
\textbf{Benchmark} & \textbf{$|$Vars$|$} & \textbf{$|$Tseitin Vars$|$} & \textbf{$|$d-DNNF$|$} & \textbf{$|$d-DNNF+p$|$} & \textbf{$|$d-DNNF+t$|$} \\
\hline
\endhead
\hline
\multicolumn{6}{r}{\textit{Continued on next page}} \\
\endfoot
\hline
\endlastfoot
card\_drawing\_10.pl & 13766 & 11775 (86\%) & 111 937 & 27 516 (25\%) & 641 (1\%) \\
biased\_coin\_toss1\_100.pl & 502 & 300 (60\%) & 12 286 & 10 498 (85\%) & 400 (3\%) \\
biased\_coin\_toss3\_100\_100\_f.pl & 601 & 398 (66\%) & 17 739 & 9 107 (51\%) & 1 647 (9\%) \\
na/newhampshire0 & 646 & 407 (63\%) & 44 326 & 9 570 (22\%) & 2 918 (7\%) \\
tree\_10\_0 & 96 & 67 (70\%) & 414 & 136 (33\%) & 53 (13\%) \\
tree\_70\_1 & 2394 & 2042 (85\%) & 429 012 & 65 235 (15\%) & 27 791 (6\%) \\
lp2sat/smokers\_8\_7 & 1897 & 1745 (92\%) & 186 150 857 & 37 841 795 (20\%) & 16 283 851 (9\%) \\
tree\_90\_1 & 3000 & 2548 (85\%) & 579 643 & 92 559 (16\%) & 41 466 (7\%) \\
tree\_20\_1 & 646 & 544 (84\%) & 26 002 & 4 630 (18\%) & 2 104 (8\%) \\
tree\_100\_1 & 3332 & 2830 (85\%) & 729 133 & 119 756 (16\%) & 55 687 (8\%) \\
lp2sat/smokers\_7\_6 & 1443 & 1324 (92\%) & 15 342 276 & 2 909 745 (19\%) & 1 361 010 (9\%) \\
tree\_80\_1 & 2737 & 2335 (85\%) & 467 568 & 76 424 (16\%) & 35 822 (8\%) \\
tree\_40\_1 & 1362 & 1160 (85\%) & 128 369 & 21 167 (16\%) & 10 060 (8\%) \\
tree\_60\_1 & 2070 & 1768 (85\%) & 243 367 & 39 684 (16\%) & 19 108 (8\%) \\
tree\_50\_1 & 1729 & 1477 (85\%) & 170 358 & 28 207 (17\%) & 13 774 (8\%) \\
na/arkansas0\_2 & 765 & 486 (64\%) & 61 091 & 17 342 (28\%) & 8 542 (14\%) \\
smokers/smokers\_8\_7 & 736 & 656 (89\%) & 6 658 551 & 1 914 221 (29\%) & 956 492 (14\%) \\
gh\_6 & 74 & 53 (72\%) & 2 072 & 786 (38\%) & 404 (19\%) \\
tree\_10\_1 & 251 & 199 (79\%) & 10 686 & 2 835 (27\%) & 1 458 (14\%) \\
tree\_30\_1 & 1022 & 870 (85\%) & 80 661 & 14 296 (18\%) & 7 456 (9\%) \\
tree\_30\_2 & 1858 & 1644 (88\%) & 3 451 907 & 459 994 (13\%) & 244 253 (7\%) \\
smokers/smokers\_7\_6 & 518 & 455 (88\%) & 1 444 858 & 434 244 (30\%) & 234 901 (16\%) \\
lp2sat/smokers\_11\_4 & 1629 & 1524 (94\%) & 8 791 418 & 919 749 (10\%) & 502 144 (6\%) \\
tree\_50\_2 & 3128 & 2774 (89\%) & 21 929 055 & 2 617 220 (12\%) & 1 428 938 (7\%) \\
tree\_60\_2 & 3740 & 3316 (89\%) & 34 091 435 & 3 920 293 (11\%) & 2 163 597 (6\%) \\
na/mexicocity0 & 333 & 204 (61\%) & 13 799 & 3 385 (25\%) & 1 873 (14\%) \\
tree\_40\_2 & 2413 & 2129 (88\%) & 9 791 517 & 1 153 522 (12\%) & 638 534 (7\%) \\
lp2sat/smokers\_10\_5 & 1193 & 1095 (92\%) & 6 053 347 & 778 090 (13\%) & 434 753 (7\%) \\
gh\_7 & 104 & 76 (73\%) & 6 314 & 2 387 (38\%) & 1 358 (22\%) \\
gh\_5 & 49 & 34 (69\%) & 852 & 331 (39\%) & 193 (23\%) \\
gh\_9 & 179 & 134 (75\%) & 46 021 & 17 176 (37\%) & 10 107 (22\%) \\
gh\_8 & 139 & 103 (74\%) & 17 269 & 6 244 (36\%) & 3 734 (22\%) \\
tree\_20\_2 & 1182 & 1038 (88\%) & 1 026 192 & 144 424 (14\%) & 86 798 (8\%) \\
gh\_10 & 224 & 169 (75\%) & 154 798 & 57 898 (37\%) & 35 186 (23\%) \\
gh\_15 & 524 & 404 (77\%) & 47 061 892 & 17 210 730 (37\%) & 10 551 421 (22\%) \\
gh\_14 & 454 & 349 (77\%) & 14 556 112 & 5 343 502 (37\%) & 3 277 196 (23\%) \\
lp2sat/smokers\_9\_5 & 867 & 792 (91\%) & 594 712 & 68 184 (11\%) & 42 006 (7\%) \\
tree\_10\_2 & 578 & 504 (87\%) & 268 341 & 42 249 (16\%) & 26 092 (10\%) \\
tree\_20\_3 & 1940 & 1754 (90\%) & 58 459 455 & 6 399 205 (11\%) & 3 988 155 (7\%) \\
lp2sat/smokers\_11\_5 & 1806 & 1693 (94\%) & 26 030 467 & 2 164 333 (8\%) & 1 352 240 (5\%) \\
gh\_11 & 274 & 208 (76\%) & 480 258 & 177 134 (37\%) & 110 946 (23\%) \\
lp2sat/smokers\_13\_3 & 1838 & 1719 (94\%) & 71 715 154 & 5 621 907 (8\%) & 3 544 183 (5\%) \\
lp2sat/smokers\_10\_4 & 1069 & 979 (92\%) & 2 311 217 & 232 632 (10\%) & 146 807 (6\%) \\
gh\_12 & 329 & 251 (76\%) & 1 267 316 & 453 895 (36\%) & 291 926 (23\%) \\
lp2sat/smokers\_12\_3 & 1915 & 1795 (94\%) & 34 645 012 & 2 433 502 (7\%) & 1 569 165 (5\%) \\
gh\_16 & 740 & 604 (82\%) & 83 689 609 & 22 402 902 (27\%) & 14 500 697 (17\%) \\
smokers/smokers\_14\_5 & 798 & 700 (88\%) & 34 464 202 & 9 089 030 (26\%) & 5 902 348 (17\%) \\
gh\_13 & 389 & 298 (77\%) & 3 651 005 & 1 284 063 (35\%) & 835 151 (23\%) \\
gh\_4 & 29 & 19 (66\%) & 239 & 103 (43\%) & 67 (28\%) \\
lp2sat/smokers\_7\_4 & 456 & 411 (90\%) & 16 874 & 1 731 (10\%) & 1 129 (7\%) \\
smokers/smokers\_13\_5 & 681 & 597 (88\%) & 15 737 234 & 4 210 361 (27\%) & 2 748 694 (17\%) \\
lp2sat/smokers\_9\_4 & 867 & 792 (91\%) & 487 149 & 45 485 (9\%) & 29 864 (6\%) \\
lp2sat/smokers\_14\_3 & 2113 & 1979 (94\%) & 72 449 868 & 5 608 578 (8\%) & 3 682 952 (5\%) \\
lp2sat/smokers\_10\_3 & 1298 & 1212 (93\%) & 1 862 646 & 158 308 (8\%) & 104 051 (6\%) \\
tree\_10\_4 & 1304 & 1186 (91\%) & 129 094 229 & 14 303 711 (11\%) & 9 404 966 (7\%) \\
lp2sat/smokers\_8\_2 & 352 & 304 (86\%) & 9 926 & 1 161 (12\%) & 765 (8\%) \\
lp2sat/smokers\_9\_3 & 867 & 792 (91\%) & 315 460 & 29 518 (9\%) & 19 465 (6\%) \\
lp2sat/smokers\_6\_2 & 277 & 239 (86\%) & 12 435 & 1 639 (13\%) & 1 082 (9\%) \\
lp2sat/smokers\_7\_3 & 456 & 411 (90\%) & 14 773 & 1 550 (10\%) & 1 032 (7\%) \\
lp2sat/smokers\_11\_3 & 1396 & 1303 (93\%) & 1 691 231 & 139 915 (8\%) & 93 236 (6\%) \\
lp2sat/smokers\_14\_2 & 1769 & 1651 (93\%) & 10 374 869 & 777 505 (7\%) & 520 046 (5\%) \\
lp2sat/tree\_10\_0 & 755 & 709 (94\%) & 2 831 & 357 (13\%) & 239 (8\%) \\
tree\_10\_3 & 933 & 837 (90\%) & 4 717 504 & 603 456 (13\%) & 405 405 (9\%) \\
smokers/smokers\_13\_4 & 571 & 491 (86\%) & 2 183 708 & 603 412 (28\%) & 408 403 (19\%) \\
lp2sat/smokers\_8\_3 & 597 & 541 (91\%) & 106 363 & 9 112 (9\%) & 6 203 (6\%) \\
lp2sat/smokers\_6\_3 & 233 & 199 (85\%) & 4 239 & 635 (15\%) & 434 (10\%) \\
lp2sat/smokers\_8\_5 & 595 & 539 (91\%) & 40 198 & 4 163 (10\%) & 2 853 (7\%) \\
smokers/smokers\_12\_5 & 518 & 442 (85\%) & 1 924 863 & 585 345 (30\%) & 401 907 (21\%) \\
lp2sat/smokers\_13\_2 & 1518 & 1415 (93\%) & 198 518 & 15 619 (8\%) & 10 744 (5\%) \\
lp2sat/smokers\_10\_2 & 1050 & 976 (93\%) & 317 339 & 26 413 (8\%) & 18 335 (6\%) \\
smokers/smokers\_15\_4 & 591 & 501 (85\%) & 2 397 884 & 684 620 (29\%) & 482 867 (20\%) \\
smokers/smokers\_9\_3 & 203 & 161 (79\%) & 9 955 & 3 535 (36\%) & 2 496 (25\%) \\
lp2sat/smokers\_15\_2 & 2046 & 1913 (93\%) & 29 694 411 & 2 070 764 (7\%) & 1 465 241 (5\%) \\
lp2sat/smokers\_8\_4 & 672 & 612 (91\%) & 56 444 & 5 352 (9\%) & 3 795 (7\%) \\
smokers/smokers\_12\_4 & 498 & 428 (86\%) & 1 424 575 & 387 035 (27\%) & 275 147 (19\%) \\
lp2sat/smokers\_12\_2 & 1315 & 1223 (93\%) & 991 917 & 76 021 (8\%) & 54 240 (5\%) \\
lp2sat/smokers\_7\_2 & 392 & 351 (90\%) & 6 116 & 646 (11\%) & 462 (8\%) \\
smokers/smokers\_12\_3 & 528 & 462 (88\%) & 639 673 & 148 377 (23\%) & 107 119 (17\%) \\
lp2sat/smokers\_11\_2 & 1144 & 1063 (93\%) & 248 139 & 17 835 (7\%) & 12 902 (5\%) \\
smokers/smokers\_9\_2 & 163 & 125 (77\%) & 3 928 & 1 607 (41\%) & 1 173 (30\%) \\
lp2sat/smokers\_9\_2 & 750 & 683 (91\%) & 72 120 & 7 043 (10\%) & 5 146 (7\%) \\
smokers/smokers\_10\_5 & 322 & 268 (83\%) & 93 962 & 28 783 (31\%) & 21 032 (22\%) \\
05\_smokers\_10\_2\_f.pl & 402 & 339 (84\%) & 1 635 018 & 197 490 (12\%) & 144 333 (9\%) \\
smokers/smokers\_7\_3 & 103 & 77 (75\%) & 1 107 & 550 (50\%) & 404 (36\%) \\
smokers/smokers\_7\_5 & 83 & 61 (73\%) & 385 & 136 (35\%) & 100 (26\%) \\
smokers/smokers\_9\_5 & 211 & 169 (80\%) & 14 509 & 5 168 (36\%) & 3 802 (26\%) \\
smokers/smokers\_13\_3 & 423 & 357 (84\%) & 201 327 & 56 048 (28\%) & 41 440 (21\%) \\
smokers/smokers\_11\_5 & 395 & 333 (84\%) & 149 664 & 42 562 (28\%) & 31 516 (21\%) \\
smokers/smokers\_15\_3 & 615 & 531 (86\%) & 1 841 428 & 444 181 (24\%) & 329 683 (18\%) \\
lp2sat/smokers\_6\_4 & 187 & 157 (84\%) & 1 395 & 214 (15\%) & 159 (11\%) \\
smokers/smokers\_9\_4 & 211 & 169 (80\%) & 11 594 & 3 861 (33\%) & 2 869 (25\%) \\
smokers/smokers\_8\_5 & 132 & 100 (76\%) & 1 969 & 744 (38\%) & 554 (28\%) \\
smokers/smokers\_14\_4 & 614 & 528 (86\%) & 7 202 247 & 1 859 261 (26\%) & 1 389 939 (19\%) \\
smokers/smokers\_10\_4 & 266 & 216 (81\%) & 42 602 & 14 285 (34\%) & 10 682 (25\%) \\
smokers/smokers\_14\_3 & 492 & 418 (85\%) & 323 550 & 86 608 (27\%) & 64 880 (20\%) \\
smokers/smokers\_8\_4 & 142 & 108 (76\%) & 2 020 & 804 (40\%) & 604 (30\%) \\
lp2sat/smokers\_5\_2 & 133 & 110 (83\%) & 703 & 113 (16\%) & 85 (12\%) \\
smokers/smokers\_11\_4 & 339 & 281 (83\%) & 154 297 & 48 349 (31\%) & 36 506 (24\%) \\
smokers/smokers\_11\_2 & 193 & 147 (76\%) & 3 193 & 1 216 (38\%) & 920 (29\%) \\
smokers/smokers\_15\_2 & 485 & 411 (85\%) & 440 143 & 110 887 (25\%) & 84 154 (19\%) \\
smokers/smokers\_10\_3 & 270 & 222 (82\%) & 23 305 & 6 999 (30\%) & 5 339 (23\%) \\
smokers/smokers\_13\_2 & 299 & 241 (81\%) & 6 351 & 1 891 (30\%) & 1 443 (23\%) \\
lp2sat/smokers\_5\_3 & 133 & 110 (83\%) & 719 & 109 (15\%) & 84 (12\%) \\
smokers/smokers\_8\_3 & 156 & 124 (79\%) & 3 020 & 958 (32\%) & 739 (24\%) \\
lp2sat/tree\_50\_1 & 12662 & 12211 (96\%) & 24 898 141 & 1 358 634 (5\%) & 1 048 205 (4\%) \\
lp2sat/tree\_60\_1 & 15382 & 14841 (96\%) & 45 619 048 & 2 103 378 (5\%) & 1 623 763 (4\%) \\
smokers/smokers\_14\_2 & 499 & 433 (87\%) & 362 333 & 71 563 (20\%) & 55 476 (15\%) \\
lp2sat/tree\_20\_1 & 3822 & 3641 (95\%) & 1 253 443 & 96 084 (8\%) & 74 498 (6\%) \\
smokers/smokers\_7\_4 & 103 & 77 (75\%) & 812 & 308 (38\%) & 239 (29\%) \\
lp2sat/tree\_40\_1 & 9990 & 9629 (96\%) & 12 153 003 & 727 574 (6\%) & 564 705 (5\%) \\
smokers/smokers\_12\_2 & 258 & 206 (80\%) & 15 773 & 5 068 (32\%) & 3 948 (25\%) \\
smokers/smokers\_11\_3 & 297 & 245 (82\%) & 21 180 & 6 123 (29\%) & 4 772 (23\%) \\
smokers/smokers\_6\_3 & 74 & 54 (73\%) & 368 & 151 (41\%) & 118 (32\%) \\
smokers/smokers\_10\_2 & 220 & 178 (81\%) & 13 440 & 4 111 (31\%) & 3 222 (24\%) \\
smokers/smokers\_6\_2 & 88 & 66 (75\%) & 965 & 371 (38\%) & 291 (30\%) \\
smokers/smokers\_7\_2 & 89 & 65 (73\%) & 447 & 187 (42\%) & 148 (33\%) \\
lp2sat/tree\_20\_2 & 5832 & 5568 (95\%) & 137 062 107 & 9 841 020 (7\%) & 7 807 204 (6\%) \\
lp2sat/tree\_30\_1 & 6322 & 6051 (96\%) & 6 068 033 & 408 990 (7\%) & 324 796 (5\%) \\
eu/no2 & 416 & 237 (57\%) & 80 572 & 16 140 (20\%) & 12 895 (16\%) \\
lp2sat/smokers\_4\_2 & 53 & 37 (70\%) & 195 & 46 (24\%) & 37 (19\%) \\
lp2sat/tree\_10\_1 & 1552 & 1461 (94\%) & 204 505 & 19 843 (10\%) & 15 987 (8\%) \\
smokers/smokers\_5\_2 & 45 & 31 (69\%) & 146 & 62 (42\%) & 50 (34\%) \\
gh\_3 & 14 & 8 (57\%) & 67 & 31 (46\%) & 25 (37\%) \\
lp2sat/smokers\_7\_5 & 326 & 289 (89\%) & 2 365 & 277 (12\%) & 224 (9\%) \\
lp2sat/tree\_10\_3 & 3131 & 2954 (94\%) & 155 421 713 & 14 425 885 (9\%) & 11 685 815 (8\%) \\
lp2sat/tree\_10\_2 & 2275 & 2141 (94\%) & 4 060 745 & 401 091 (10\%) & 325 151 (8\%) \\
na/illinois1\_4 & 436 & 271 (62\%) & 210 624 & 60 538 (29\%) & 49 219 (23\%) \\
smokers/smokers\_5\_3 & 45 & 31 (69\%) & 140 & 59 (42\%) & 48 (34\%) \\
smokers/smokers\_8\_2 & 110 & 82 (75\%) & 852 & 334 (39\%) & 272 (32\%) \\
na/illinois1\_3 & 455 & 290 (64\%) & 172 524 & 46 592 (27\%) & 38 062 (22\%) \\
na/nuevoleon0\_3 & 485 & 321 (66\%) & 222 888 & 63 957 (29\%) & 53 252 (24\%) \\
eu/hr0 & 401 & 255 (64\%) & 349 915 & 95 158 (27\%) & 79 612 (23\%) \\
smokers/smokers\_6\_4 & 66 & 48 (73\%) & 365 & 140 (38\%) & 118 (32\%) \\
na/westvirginia0 & 548 & 404 (74\%) & 688 822 & 125 195 (18\%) & 105 649 (15\%) \\
smokers/smokers\_4\_2 & 30 & 20 (67\%) & 73 & 34 (47\%) & 29 (40\%) \\
05\_smokers\_10\_f.pl & 229 & 168 (73\%) & 76 013 & 15 081 (20\%) & 12 952 (17\%) \\
na/nuevoleon0\_2 & 480 & 317 (66\%) & 155 228 & 41 833 (27\%) & 35 989 (23\%) \\
na/guanajuato0\_4 & 581 & 363 (62\%) & 243 381 & 71 368 (29\%) & 61 803 (25\%) \\
na/newyork1 & 382 & 255 (67\%) & 267 893 & 60 782 (23\%) & 52 819 (20\%) \\
05\_smokers\_20\_f.pl & 156 & 52 (33\%) & 1 611 & 958 (59\%) & 845 (52\%) \\
na/arkansas0\_3 & 767 & 488 (64\%) & 347 875 & 86 564 (25\%) & 76 495 (22\%) \\
na/arkansas0\_4 & 824 & 544 (66\%) & 749 924 & 199 067 (27\%) & 177 464 (24\%) \\
na/arkansas0\_1 & 869 & 590 (68\%) & 997 520 & 260 394 (26\%) & 232 210 (23\%) \\
na/nuevoleon0\_1 & 669 & 506 (76\%) & 11 881 970 & 1 508 787 (13\%) & 1 359 634 (11\%) \\
na/guanajuato0\_2 & 615 & 397 (65\%) & 703 710 & 182 359 (26\%) & 164 487 (23\%) \\
na/minnesota1 & 532 & 392 (74\%) & 1 317 519 & 261 319 (20\%) & 235 729 (18\%) \\
na/illinois1\_2 & 450 & 285 (63\%) & 334 939 & 86 464 (26\%) & 78 538 (23\%) \\
na/sanluispotosi0 & 486 & 328 (67\%) & 858 929 & 227 773 (27\%) & 207 698 (24\%) \\
na/guanajuato0\_1 & 701 & 483 (69\%) & 2 864 815 & 566 413 (20\%) & 523 804 (18\%) \\
na/guanajuato0\_3 & 613 & 395 (64\%) & 1 012 685 & 240 711 (24\%) & 223 486 (22\%) \\
eu/ru1 & 892 & 676 (76\%) & 15 835 947 & 2 217 816 (14\%) & 2 077 827 (13\%) \\
na/bajacalifornia0 & 537 & 412 (77\%) & 2 318 894 & 265 241 (11\%) & 248 967 (11\%) \\
na/illinois1\_1 & 530 & 365 (69\%) & 876 888 & 199 911 (23\%) & 188 925 (22\%) \\
na/chihuahua0\_1 & 641 & 478 (75\%) & 47 630 040 & 6 990 905 (15\%) & 6 744 530 (14\%) \\
na/chihuahua0\_2 & 633 & 470 (74\%) & 36 016 817 & 5 836 703 (16\%) & 5 716 480 (16\%) \\
gnb\_20 & 400 & 208 (52\%) & 6 058 & 1 485 (25\%) & 1 485 (25\%) \\
gnb\_50 & 2500 & 1273 (51\%) & 144 700 & 26 511 (18\%) & 26 511 (18\%) \\
gnb\_60 & 3600 & 1828 (51\%) & 298 756 & 50 815 (17\%) & 50 815 (17\%) \\
gnb\_90 & 8100 & 4093 (51\%) & 1 623 372 & 251 742 (16\%) & 251 742 (16\%) \\
gnb\_30 & 900 & 463 (51\%) & 22 798 & 4 864 (21\%) & 4 864 (21\%) \\
gnb\_40 & 1600 & 818 (51\%) & 64 235 & 12 603 (20\%) & 12 603 (20\%) \\
gnb\_70 & 4900 & 2483 (51\%) & 543 578 & 88 635 (16\%) & 88 635 (16\%) \\
gnb\_80 & 6400 & 3238 (51\%) & 982 006 & 155 952 (16\%) & 155 952 (16\%) \\
smokers/smokers\_6\_5 & 30 & 18 (60\%) & 47 & 23 (49\%) & 23 (49\%) \\
smokers/smokers\_4\_3 & 20 & 12 (60\%) & 31 & 15 (48\%) & 15 (48\%) \\
smokers/smokers\_5\_4 & 25 & 15 (60\%) & 39 & 19 (49\%) & 19 (49\%) \\
smokers/smokers\_3\_2 & 15 & 9 (60\%) & 23 & 11 (48\%) & 11 (48\%) \\
gh\_2 & 4 & 1 (25\%) & 8 & 6 (75\%) & 6 (75\%) \\
lp2sat/smokers\_4\_4 & 9 & 5 (56\%) & 9 & 3 (33\%) & 3 (33\%) \\
lp2sat/smokers\_4\_5 & 9 & 5 (56\%) & 9 & 3 (33\%) & 3 (33\%) \\
lp2sat/smokers\_4\_3 & 21 & 9 (43\%) & 32 & 19 (59\%) & 19 (59\%) \\
lp2sat/smokers\_6\_5 & 31 & 13 (42\%) & 48 & 29 (60\%) & 29 (60\%) \\
lp2sat/smokers\_3\_4 & 7 & 4 (57\%) & 7 & 2 (29\%) & 2 (29\%) \\
lp2sat/smokers\_3\_3 & 7 & 4 (57\%) & 7 & 2 (29\%) & 2 (29\%) \\
lp2sat/smokers\_5\_5 & 11 & 6 (55\%) & 11 & 4 (36\%) & 4 (36\%) \\
lp2sat/smokers\_5\_4 & 26 & 11 (42\%) & 40 & 24 (60\%) & 24 (60\%) \\
lp2sat/smokers\_3\_2 & 16 & 7 (44\%) & 24 & 14 (58\%) & 14 (58\%) \\
lp2sat/smokers\_7\_7 & 15 & 8 (53\%) & 15 & 6 (40\%) & 6 (40\%) \\
lp2sat/smokers\_3\_5 & 7 & 4 (57\%) & 7 & 2 (29\%) & 2 (29\%) \\
weather\_50.pl & 596 & 296 (50\%) & 7 007 & 3 507 (50\%) & 3 507 (50\%) \\
\end{longtable}

\begin{longtable}{|l|r|r|r|r|r|}
\caption{Experimental results comparing the compiled formula size ($|$d-DNNF$|$) to the size after performing existential quantification ($|$d-DNNF+p$|$) and to the size after also considering Tseitin artifacts ($|$d-DNNF+t$|$). The percentages are relative to the d-DNNF size. The percentage next to $|$Tseitin Vars$|$ indicates the relative number of tseitin variables compared to the total number of variables. The instances are those of the MCC dataset.}\label{resultstable2}\\
\hline
\textbf{Benchmark} & \textbf{$|$Vars$|$} & \textbf{$|$Tseitin Vars$|$} & \textbf{$|$d-DNNF$|$} & \textbf{$|$d-DNNF+p$|$} & \textbf{$|$d-DNNF+t$|$} \\
\hline
\endfirsthead
\multicolumn{6}{c}%
{\tablename\ \thetable\ -- \textit{Continued from previous page}} \\
\hline
\textbf{Benchmark} & \textbf{$|$Vars$|$} & \textbf{$|$Tseitin Vars$|$} & \textbf{$|$d-DNNF$|$} & \textbf{$|$d-DNNF+p$|$} & \textbf{$|$d-DNNF+t$|$} \\
\hline
\endhead
\hline
\multicolumn{6}{r}{\textit{Continued on next page}} \\
\endfoot
\hline
\endlastfoot
MCC22\_wpu2\_005 & 100 & 50 (50\%) & 2 776 738 & 1 761 751 (63\%) & 1 (0\%) \\
MCC22\_wpu2\_007 & 140 & 70 (50\%) & 63 959 & 36 835 (58\%) & 6 916 (11\%) \\
MCC22\_wpu2\_013 & 100 & 50 (50\%) & 66 763 & 42 193 (63\%) & 10 838 (16\%) \\
MCC22\_wpu2\_009 & 140 & 70 (50\%) & 32 118 & 18 846 (59\%) & 7 538 (23\%) \\
MCC22\_pu1\_007 & 200 & 100 (50\%) & 687 & 328 (48\%) & 162 (24\%) \\
MCC22\_wpu2\_011 & 200 & 100 (50\%) & 7 176 802 & 5 038 851 (70\%) & 3 203 903 (45\%) \\
MCC22\_pu1\_011 & 120 & 60 (50\%) & 57 772 & 39 137 (68\%) & 27 578 (48\%) \\
MCC22\_pu1\_015 & 200 & 100 (50\%) & 239 & 119 (50\%) & 87 (36\%) \\
MCC22\_pu1\_021 & 586 & 388 (66\%) & 27 169 & 9 286 (34\%) & 6 822 (25\%) \\
MCC22\_wpu2\_093 & 1736 & 1547 (89\%) & 398 357 & 36 790 (9\%) & 28 177 (7\%) \\
MCC22\_pu1\_051 & 1060 & 735 (69\%) & 295 814 & 109 243 (37\%) & 83 723 (28\%) \\
MCC22\_pu1\_037 & 781 & 583 (75\%) & 228 199 & 82 992 (36\%) & 64 723 (28\%) \\
MCC22\_pu1\_055 & 1332 & 908 (68\%) & 874 535 & 283 090 (32\%) & 227 330 (26\%) \\
MCC23\_pr1\_066 & 2297 & 2106 (92\%) & 9 155 265 & 783 912 (9\%) & 638 138 (7\%) \\
MCC23\_pr1\_068 & 14948 & 14266 (95\%) & 1 441 086 & 43 472 (3\%) & 37 287 (3\%) \\
MCC23\_pu1\_043 & 1985 & 1280 (64\%) & 59 793 & 12 715 (21\%) & 11 215 (19\%) \\
MCC22\_pu1\_031 & 777 & 507 (65\%) & 807 597 & 234 294 (29\%) & 210 472 (26\%) \\
MCC23\_wpu2\_009 & 9532 & 8643 (91\%) & 14 746 443 & 1 223 890 (8\%) & 1 126 578 (8\%) \\
MCC22\_wpu2\_049 & 795 & 507 (64\%) & 1 521 468 & 480 206 (32\%) & 443 030 (29\%) \\
MCC23\_pu1\_037 & 2811 & 1974 (70\%) & 74 120 & 19 741 (27\%) & 19 593 (26\%) \\
MCC22\_pu1\_079 & 1548 & 913 (59\%) & 32 619 & 15 316 (47\%) & 15 211 (47\%) \\
MCC23\_wpu2\_023 & 3012 & 2089 (69\%) & 77 431 & 22 069 (29\%) & 21 921 (28\%) \\
MCC23\_wpr2\_036 & 2708 & 2014 (74\%) & 90 758 & 28 828 (32\%) & 28 714 (32\%) \\
MCC23\_wpu2\_035 & 2374 & 1447 (61\%) & 72 406 & 34 609 (48\%) & 34 504 (48\%) \\
MCC22\_pu1\_121 & 9386 & 2828 (30\%) & 36 720 080 & 26 300 884 (72\%) & 26 286 822 (72\%) \\
MCC22\_pu1\_149 & 17849 & 8694 (49\%) & 297 569 & 157 314 (53\%) & 157 250 (53\%) \\
MCC22\_wpu2\_149 & 17918 & 8694 (49\%) & 299 748 & 161 117 (54\%) & 161 053 (54\%) \\
MCC22\_wpu2\_103 & 586 & 243 (41\%) & 3 481 642 & 2 210 014 (63\%) & 2 210 014 (63\%) \\
MCC23\_pu1\_085 & 892 & 243 (27\%) & 1 916 976 & 1 425 391 (74\%) & 1 425 391 (74\%) \\
MCC22\_pu1\_099 & 1843 & 968 (53\%) & 76 554 867 & 33 765 798 (44\%) & 33 765 798 (44\%) \\
MCC23\_pu1\_027 & 1916 & 1134 (59\%) & 64 127 & 30 507 (48\%) & 30 507 (48\%) \\
MCC22\_pu1\_093 & 2065 & 1060 (51\%) & 566 809 & 338 489 (60\%) & 338 489 (60\%) \\
MCC23\_pu1\_031 & 2649 & 1905 (72\%) & 46 726 & 14 542 (31\%) & 14 542 (31\%) \\
MCC23\_pr1\_090 & 2687 & 968 (36\%) & 82 680 156 & 50 593 787 (61\%) & 50 593 787 (61\%) \\
MCC22\_wpu2\_101 & 2815 & 1060 (38\%) & 790 369 & 548 570 (69\%) & 548 570 (69\%) \\
MCC23\_wpr2\_032 & 2646 & 1985 (75\%) & 72 318 & 22 277 (31\%) & 22 277 (31\%) \\
MCC22\_pu1\_107 & 3045 & 1528 (50\%) & 13 606 870 & 8 030 943 (59\%) & 8 030 943 (59\%) \\
MCC23\_pr1\_044 & 3383 & 2347 (69\%) & 92 416 & 32 825 (36\%) & 32 825 (36\%) \\
MCC23\_pr1\_096 & 4131 & 1528 (37\%) & 21 409 219 & 14 949 035 (70\%) & 14 949 035 (70\%) \\
MCC23\_pu1\_007 & 3727 & 3234 (87\%) & 126 623 & 15 142 (12\%) & 15 142 (12\%) \\
MCC23\_pu1\_049 & 3754 & 3317 (88\%) & 1 301 455 & 130 565 (10\%) & 130 565 (10\%) \\
MCC23\_pu1\_073 & 6139 & 5461 (89\%) & 2 437 952 & 234 861 (10\%) & 234 861 (10\%) \\
MCC23\_wpu2\_033 & 6772 & 6012 (89\%) & 691 193 & 69 937 (10\%) & 69 937 (10\%) \\
MCC23\_pr1\_008 & 6856 & 6159 (90\%) & 201 009 & 20 037 (10\%) & 20 037 (10\%) \\
MCC22\_pu1\_133 & 9638 & 6358 (66\%) & 16 099 & 5 725 (36\%) & 5 725 (36\%) \\
MCC23\_pu1\_009 & 8978 & 8124 (90\%) & 262 046 & 24 516 (9\%) & 24 516 (9\%) \\
MCC23\_pu1\_105 & 11557 & 6808 (59\%) & 3 076 699 & 1 518 536 (49\%) & 1 518 536 (49\%) \\
MCC23\_pu1\_053 & 11024 & 2757 (25\%) & 4 367 241 & 3 531 666 (81\%) & 3 531 666 (81\%) \\
MCC23\_pr1\_094 & 9797 & 8794 (90\%) & 8 783 294 & 838 762 (10\%) & 838 762 (10\%) \\
MCC23\_wpr2\_050 & 11118 & 2793 (25\%) & 2 675 743 & 2 200 837 (82\%) & 2 200 837 (82\%) \\
MCC23\_pu1\_033 & 10481 & 9438 (90\%) & 193 448 & 18 991 (10\%) & 18 991 (10\%) \\
MCC23\_wpr2\_034 & 11825 & 10621 (90\%) & 359 365 & 32 681 (9\%) & 32 681 (9\%) \\
MCC23\_pr1\_072 & 14685 & 10179 (69\%) & 388 592 & 145 863 (38\%) & 145 863 (38\%) \\
MCC23\_wpu2\_081 & 13808 & 12301 (89\%) & 630 642 & 58 587 (9\%) & 58 587 (9\%) \\
MCC23\_pr1\_110 & 16162 & 14651 (91\%) & 31 486 631 & 2 553 281 (8\%) & 2 553 281 (8\%) \\
MCC23\_pr1\_100 & 16767 & 14868 (89\%) & 14 758 376 & 1 155 455 (8\%) & 1 155 455 (8\%) \\
MCC23\_pr1\_070 & 17299 & 15727 (91\%) & 838 399 & 71 378 (9\%) & 71 378 (9\%) \\
MCC23\_pu1\_117 & 20192 & 12696 (63\%) & 2 432 088 & 1 147 968 (47\%) & 1 147 968 (47\%) \\
MCC23\_pu1\_103 & 20977 & 19095 (91\%) & 9 511 202 & 783 373 (8\%) & 783 373 (8\%) \\
MCC23\_pr1\_146 & 28701 & 28343 (99\%) & 8 119 721 & 83 212 (1\%) & 83 212 (1\%) \\
MCC23\_pu1\_099 & 23246 & 21294 (92\%) & 2 910 837 & 218 127 (7\%) & 218 127 (7\%) \\
MCC23\_pu1\_119 & 37821 & 22486 (59\%) & 17 933 041 & 9 254 462 (52\%) & 9 254 462 (52\%) \\
MCC23\_wpr2\_098 & 33446 & 30521 (91\%) & 1 497 939 & 103 126 (7\%) & 103 126 (7\%) \\
MCC23\_pu1\_133 & 44805 & 27489 (61\%) & 55 146 328 & 25 261 313 (46\%) & 25 261 313 (46\%) \\
MCC23\_pr1\_032 & 192958 & 63516 (33\%) & 2 352 179 & 1 562 547 (66\%) & 1 562 547 (66\%) \\
\end{longtable}

\end{document}